# Changing the Image Memorability: From Basic Photo Editing to GANs

Oleksii Sidorov
The Norwegian Colour and Visual Computing Laboratory, NTNU
Gjøvik, Norway
oleksiis@stud.ntnu.no

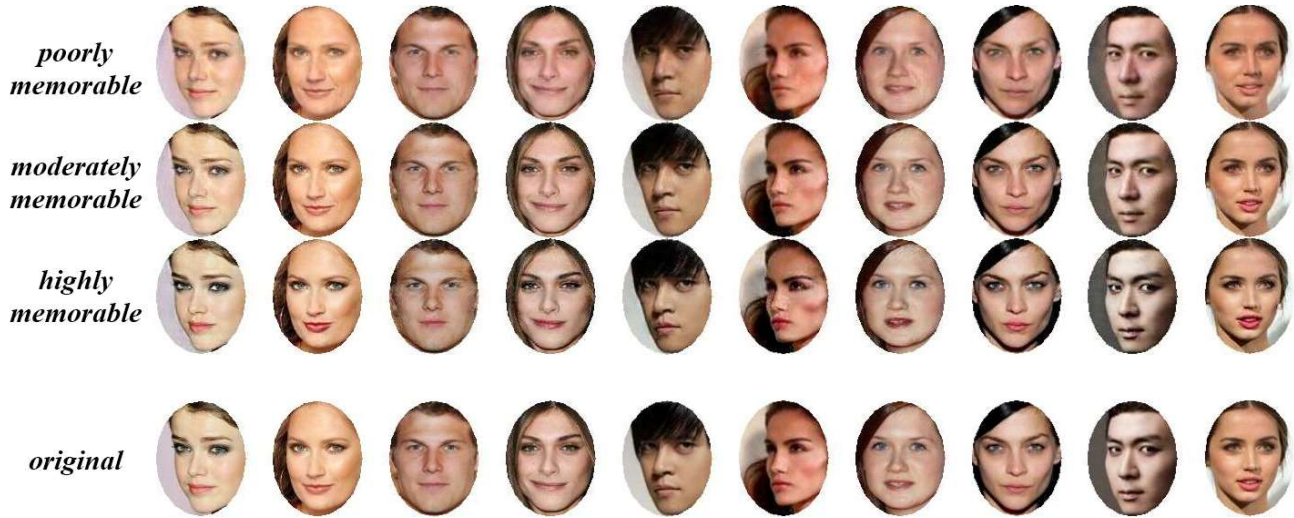

Figure 1: Modification of memorability using the proposed algorithm. All the results were generated without any human intervention. "What" and "how" to change were learned by the model from experimental data.

## Abstract

*Memorability is considered to be an important characteristic of visual content, whereas for advertisement and educational purposes it is often crucial. Despite numerous studies on understanding and predicting image memorability, there are almost no achievements in memorability modification. In this work, we study two approaches to image editing – GAN and classical image processing – and show their impact on memorability.*

*The visual features which influence memorability directly stay unknown till now, hence it is impossible to control it manually. As a solution, we let GAN learn it deeply using labeled data, and then use it for conditional generation of new images. By analogy with algorithms which edit facial attributes, we consider memorability as yet another attribute and operate with it in the same way. Obtained data is also interesting for analysis, simply because there are no real-world examples of successful change of image memorability while preserving its other attributes. We believe this may give many new answers to the question "what makes an image memorable?"*

*Apart from that we also study the influence of conventional photo-editing tools (Photoshop, Instagram, etc.) used daily by a wide audience on memorability. In this case, we start from real practical methods and study it using statistics and recent advances in memorability prediction. Photographers, designers, and advertisers will benefit from the results of this study directly.*

## 1. Introduction

Memory is a strange thing. You probably remember the photo of The Beatles crossing the road which you saw many years ago, but do not remember the poster on the storefront you saw recently. Is it only a matter of attention and context? Numerous experiments show that memorability is an intrinsic feature of an image (or a video) which is also highly consistent between different observers. So, eventually, memorability is but another parameter we can measure, describe, predict, or even change. The majority of existing studies on media memorability are dedicated to the prediction of memorability score of a given image or a video [2][4][14][21][37][41]. A few others tried to understand what actually influences media memorability [8][15][16][21], and very rare tried to modify this value [19][39]. However, for practical application exactly the last part is the most important one, while understanding and prediction are only the tools which can help to achieve it. So, this work is dedicated particularly to the ultimate goal of image memorability research – image modification.

In the first part, we show how image memorability can benefit from state-of-the-art deep learning techniques designed for controlled image-to-image translation [5][12][28]. By analogy with learning-based modification of facial attributes (such as hair color, age, smile, etc.) we modify memorability considering it as yet another facial attribute. The main requirement (and the main obstacle) of such approach is the need for a relatively large set of labeled data. To resolve this problem, we created synthetic labels for 200k images of faces from CelebA [31] dataset, using predictor trained on 2,222 images with scores collected from real observers in the experiment [1]. The difference in the images generated by the given algorithm can be seen with the naked eye (Fig. 1). Moreover, there are no analogs of real-world data which could provide different memorability scores for the same objects or scenes. This fact will allow to reverse-engineer the output in order to analyze how exactly the image was changed and which features played the most significant role from the perspective of GAN itself. Such visualization also shows how visual features learned by the model are mapped to memorability scores. We believe it will produce many insights for other researchers.

In the second part, we do not try to identify abstract mathematical or psychological features which may influence image memorability (such as different statistical values, distributions, clusters or maps) simply because photographers, designers and other artists who actually create media will not be able to use it in their work. Instead, we start from the end and study how conventional photo-editing tools which people use daily influence image memorability. Such tools include but are not limited to contrast enhancement, sharpening, filtering, background blurring, color temperature, and others. Are they able to increase memorability? Are they safe to use in order not to decrease it? Do they influence all the images in the same way? These and other questions we are trying to answer. The most accurate way to do it would be to process the data using given tools, collect observers' responses and compare them to the original ones. But due to a large amount of data and peculiarities of memorability measurement the cost of such an experiment would be enormously high. So, as an approximation model to this "ideal" pipeline, we use state-of-the-art memorability predictor which achieves human-like consistency of memorability estimation [9]. This allowed us to experiment with more than 200,000 images and collect statistically significant data. We understand that these results are only an approximation to human responses, but considering that the predictor was trained on 45,000 images evaluated by humans, we have reasons to trust their reliability.

## 2. Related works

2.1. Image Memorability

Memorability is a versatile topic. Different types of memory, as well as dependence on context and subjective variations, do not allow to describe memorability with a single simple model. However, in the pioneer study, Isola *et al.* [14] proved experimentally that memorability can be considered an intrinsic feature of an image, that provides consistent results between different groups of observers. Isola *et al.* [14] also created the first image memorability dataset of 2,222 images with corresponding memorability scores. Both of these factors created a basis for further research. A number of following works were dedicated to understanding [3][15][16] or prediction [4][20][23] of image memorability using classical hand-crafted features, namely SIFT [32], HOG2x2 [7], semantic labels, and pixel color histograms. The following achievements of deep learning techniques for computer vision tasks were a good motivation for using them in the memorability domain. So, starting from 2015 top accuracy in memorability prediction was achieved particularly by CNN-based algorithms. Firstly, it was demonstrated independently by Khosla *et al.* [21] and Baveye *et al.* [2] how simple fine-tuning of well-known CNNs to regression task could outperform all previous prediction models. In the same work, Khosla *et al.* [21] presented a novel, much larger dataset called LaMem (60k images) which has become dominant in the field. Following works elaborated on this idea by adding additional modules, such as Image Captioning [41], Segmentation Maps [44], and Attention Maps [8][9]. The algorithm described in last-mentioned work – AMNet – is the most recent and most accurate predictor available, so, we selected it for our experiments. Apart of prediction, there are a number of highly related works dedicated to study of extrinsic conditions [3], memorability maps [20][22], biological effects [42], and video memorability [6][11][37], which can be considered as an extension to single-image memorability. Despite significant progress in memorability research, there is still no clear vision of what exactly defines image memorability. It was proven [14][16] that the semantics plays the most important role, but what if the application requires keeping the semantics unchanged, what else should we modify?

2.2. Modification of image memorability

To the best of our knowledge, there are only two works aimed at modification of image memorability. An algorithm proposed by Khosla et al. [19] showed excellent performance on modification of face photographs. Using Active Appearance Model (AAM) to represent a face in lower dimensionality, given algorithm optimizes custom cost function, which aimed at changing memorability score while penalizing change of other features. For reasons unknown, despite promising results, this study did not receive further development. Another approach authored by Siarohin et al. [39] utilize deep style transfer to change memorability score of an image. Despite the potential ability of the algorithm to change memorability, it can hardly be used for practical purposes. The images obtained in result differ significantly from the original ones in many aspects; furthermore, in the majority of cases, edited photos lose their realism.

2.3. Image generation

Recent development of Generative Adversarial Networks (GANs) [10] gave rise to a new era in synthetic image generation. GANs have shown remarkable results in various computer vision tasks such as image generation [13][18][36][45], image translation [17][24][46], video translation [43], deblurring [27], segmentation [33], super-resolution imaging [29], and face image synthesis [25][30][38]. Due to the high quality of results and possibility to control output by given condition list of the domains where GANs were successfully applied grows rapidly. A core principle behind any GAN model is a competition between two modules: a discriminator and a generator. The discriminator learns to distinguish between real and fake samples, while the generator learns to generate fake samples that are indistinguishable from real samples. Development of conditional generation allowed to use class labels, attributes, text description, or even other images [17] to control the output. A significant part of these algorithms is dedicated to facial attribute editing (that eventually may be extended to any attribute). VAE/GAN [28] represents a combination of VAE [26] and GAN [10]. During training, VAE/GAN learns the latent representation of a data. Attribute editing then performed as algebraic operations on image representation in latent space. IcGAN [35] separately trains a cGAN [34] and an encoder. After that, images can be modified in a similar way: encoding them to latent representations and using the latent vector as a condition for the cGAN generator. One of the most recent algorithms AttGAN [12] also utilizes encoder-decoder scheme but in contrast to previous algorithms removes the strict attribute-independent constraint from the latent representation, and only applies the attribute classification constraint to the generated image to guarantee the correct change of the attributes. As authors claim, it allows to achieve more accurate editing while preserving the details and whole image visually realistic. Another recent work StarGAN [5] has a feature to perform mapping among multiple domains using only a single generator and a discriminator. In comparison to AttGAN, StarGAN uses cycle consistency loss while AttGAN does not include cyclic process or cycle loss; moreover, StarGAN trains a conditional attribute transfer network and does not involve any latent representation. In our work we benefit from the remarkable performance of the described algorithms utilizing them for modification of another image attribute – memorability.

**3. GANs for memorability modification**

3.1. Design of a dataset

Naturally, learning-based methods highly depend on data. Training of generative encoder-decoder system is even more demanding to data set in comparison to typical CNN. The one dataset which is commonly used in facial attributes modification is CelebA [31]. It includes 202,599 photos of celebrities' faces annotated with 40 binary attributes, such as hair color, gender, mustache, smile, fringe, etc. The number of photos and their similarity (faces only) make this dataset well suitable for the tasks of generation and modification. However, in the case of memorability, the size of available datasets is much more limited [1][14][31]. So, the design of appropriate training dataset was an important task in this part of the work.

"10k US Adult Faces Database" [1] is the only available dataset of face photographs with related memorability data. Apart from fillers, it consists of 2,222 images annotated with memorability scores collected in a "Memory Game" experiment from human observers. Two possible approaches have been proposed initially: to use small (2k) original dataset with real scores, or a big (200k) dataset with synthetic scores. The additional issue was the format of the attributes. As mentioned before, CelebA contains a list of binary attributes (1 – present, 0 – absent), memorability score in return is a floating point number in the range from 0 to 1. This obstacle was solved by transforming the memorability scores into three labels: low, medium and high. Thresholds between these labels were found in a heuristic way as splits with different ratio of data. In the experiments we tried such splits (low – med – high correspondingly): 1/3 – 1/3 – 1/3, 0.25 – 0.5 – 0.25, 0.1 – 0.8 – 0.1. In the result, the split 10% – low, 80% – med, and 10% – high showed the best performance. So, corresponding thresholds were equal to 10th and 90th percentiles of all scores. Using 10 classes instead of 3 has been tried as well, but did not show satisfactory results.

The straightforward use of small original dataset for training GAN expectedly failed. Eventually, 2,222 images were not enough. This created a need to expand CelebA dataset with memorability information. The most optimal way to do it was to train accurate predictor to simulate human responses. This was made possible utilizing AMNet [9] – state-of-the-art memorability predictor with human-like consistency (0.677 (AMNet) against 0.68 (observers) on the Spearman's rank correlation). In the next step, the source code of AMNet was used to train it with 2,222 images from [1]. Additionally, we modified CelebA images in order to correspond to the format of training data: all the images were masked with an oval mask to hide the background, then square-cropped to bound an oval, and finally resized to 128x128 pixels size (Fig. 2). In result, having trained the predictor and acquired data in a suitable format, we were able to estimate approximate memorability scores for 200k CelebA images. Afterward, these values were transformed into 3 classes, as was mentioned before; threshold values between classes were equal to 0.44 and 0.63 (low-medium and medium-high respectively).

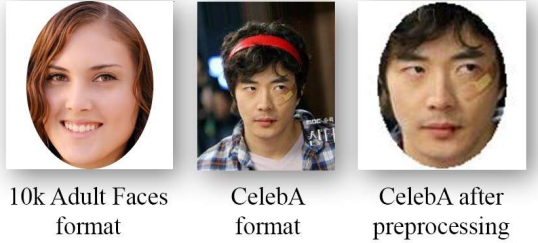

| 10k Adult Faces format | CelebA format | CelebA after preprocessing |

Figure 2: Preprocessing of CelebA dataset [31] to match the format of images with annotated memorability [1].

In conclusion, approximate scores for 200k CelebA images were synthesized using real data of 2,222 face photos. Synthesized scores are not real human responses, but the response of a deep network based on what it learned from real-human data. The distribution of predicted scores is illustrated in Fig. 3. The shape of a histogram is similar to the Gaussian distribution that demonstrates the natural-like behavior of a predictor.

### 3.2. Experiment

The described set of images annotated with memorability labels was used to train various generative algorithms. In our experiment, we used VAE/GAN [28], StarGAN [5] and AttGAN [12]. However, only AttGAN produced satisfactory output – visually distinguishable images of sufficient quality. Training was performed during 100 epochs with default parameter values specified by authors. The results generated are illustrated in Fig. 1. The difference of visual appearance of the same face with different memorability level is obvious. However, the essential question remains: did the memorability actually change? In order to estimate this, firstly we used the same AMNet predictor trained on 2,222 human-annotated images, and then validated the results using psychophysical experiment with real observers.

The result of the synthetic evaluation of 220 test images is presented in Fig. 4. The order of images is sorted according to the original memorability score (thick black line). Blue and black lines demonstrate memorability scores of the same image modified with "high" or "low" attribute added, respectively. The additional bar chart illustrates difference (Δ) between the original and modified score. It clearly shows that, with exception to a few outliers, all the scores were changed in correctly specified direction. The reader may be confused by poor results on the edges of the range. Nevertheless, we need to explain that such behavior is expected due to the specifics of data labeling. For instance, adding an attribute "high" (0.63 - 0.87) to an image with original score 0.8 will not increase it, but may, instead, decrease it to a mean score of a "high" class. To increase the memorability of an image with original score 0.8 would require the creation of another "ultra-high" class with scores above 0.8. The same is fair for the "low" class, so the adequate evaluation of results should exclude "low-low" and "high-high" translation. Table 1 describes the change of score ($\Delta_L$ and $\Delta_H$) statistically. Statistical values in the table characterize proposed algorithm very positively. In both directions, it achieves the change of memorability by 0.09 on average, with at least 95% of data being modified in the correct direction. Maximum |Δ| achieved among tested 220 images are 0.21 and 0.33 for decreasing and increasing respectively.

#### 3.2.1 Validation in psychophysical experiment

Following the procedure described in [19], we used "Memory Game" launched at Amazon Mechanical Turk to gather human responses. We used a subset of 50 modified images as targets, while other modifications were mixed with CelebA images and were used as fillers. Observers were split into two groups, so that they saw either version with increased or decreased memorability, but never both. On average, 60 responses were received for each image.

The results (Fig. 5, Table 1) demonstrate the successful change of memorability for most of the images. In case of 92% of the images, the difference between "increased" and "decreased" scores is positive, in comparison to 74% in [19] (although we used much smaller set). The comparison with original values ($\Delta_L$ and $\Delta_H$) shows similar results to the ones predicted by AMNet, however with much higher variance.

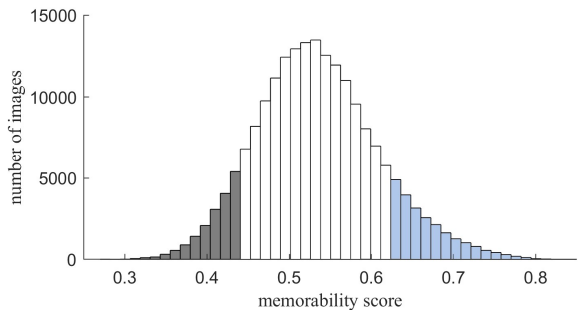

Figure 3: A distribution of memorability scores predicted for CelebA images. Colors correspond to the following separation on "low," "medium" and "high" classes (10% – 80% – 10%).

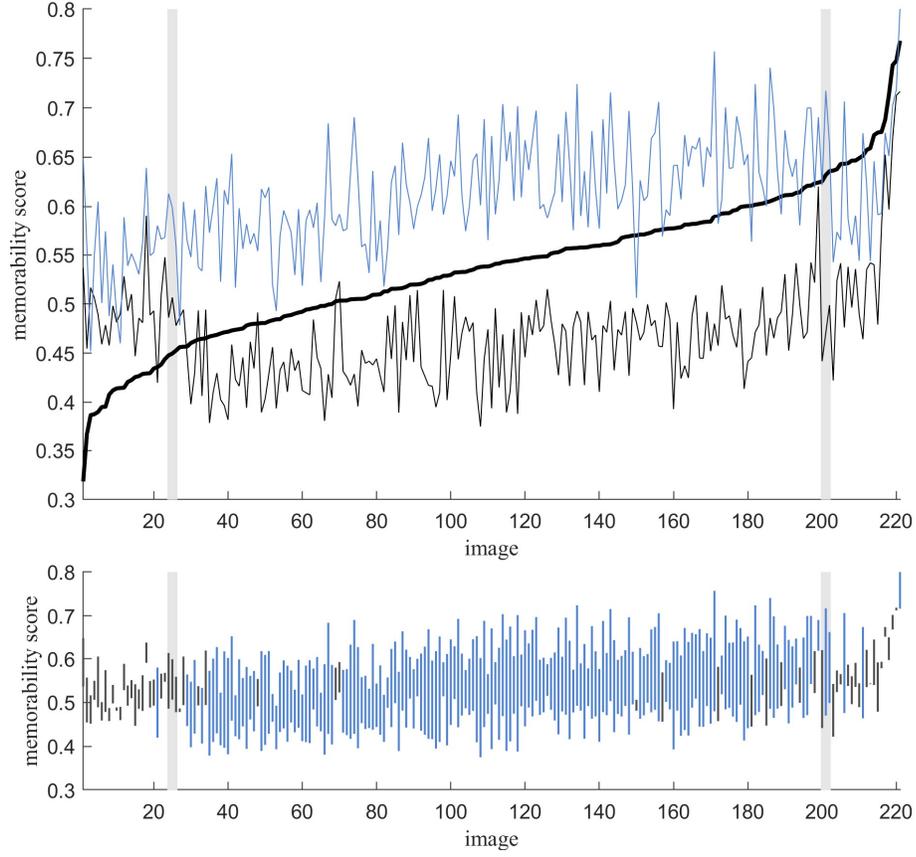

Figure 4: Memorability of images modified using AttGAN predicted by AMNet. Top: plots of memorability scores of 220 tested images; thick black line – original images, blue line – attribute "high" added, black line – attribute "low" added. Bottom: memorability change range chart, longer bars correspond to better results; blue – both modifications are correct, black – either "high" is lower than original or "low" is higher than original. Thick vertical lines indicate margins of "low," "medium" and "high" memorability classes. Images are sorted by original memorability score. Zoom is required.

TABLE 1. Change of memorability score (Δ) by AttGAN.

| Direction | Prediction by AMNet [220 images] | | | | | Results from Memory Game [50 images] | | | | |
|---|---|---|---|---|---|---|---|---|---|---|
| | mean | std | median | 5th % | 95th % | mean | std | median | 5th % | 95th % |
| **Increasing ($\Delta_H$)** | 0,086 | 0,053 | 0,083 | 0,007 | 0,167 | 0,079 | 0,090 | 0,077 | -0,068 | 0,242 |
| **Decreasing ($\Delta_L$)** | -0,088 | 0,047 | -0,087 | -0,159 | -0,010 | -0,051 | 0,084 | -0,063 | -0,150 | 0,137 |

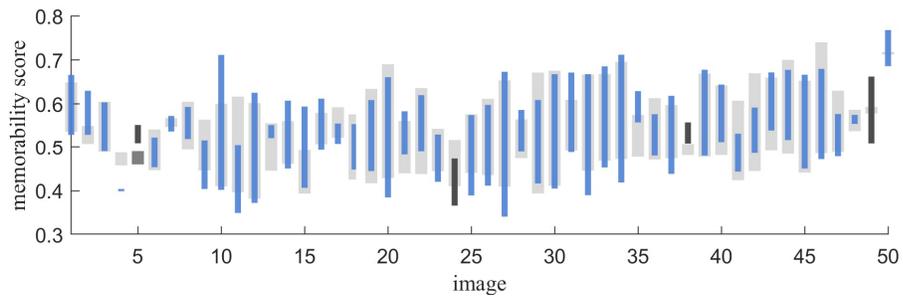

Figure 5: Memorability of images modified using AttGAN evaluated in Memory Game. Top value of the range – memorability score of the image with increased memorability, bottom value of the range – decreased. Black bars – vice versa (negative change). Thick grey bars illustrate results predicted by AMNet for a given image. Images are sorted by original memorability score. Zoom is required.

## 3.3. Analysis of generated images

More samples of generated data are available in Appendix A. Visually, images translated to "low" class look monotonic and faint, while samples modified as highly memorable look vivid and high-contrast. The naïve observers describe images with increased memorability as being sharper and clearer, while some of them also being more "exaggerated" or "pronounced" than images with decreased memorability score. Talking about facial changes, increasing the memorability score led to the following changes: makeup on women's faces became more apparent and saturated; lips became thicker and more colorful, while the width of the mouth decreased; the eyes became bigger and more distinct; the relief of the face became sharper (due to the deeper shadows) in contrast to flat relief of poorly memorable images; hairstyle, nose, and shape of the face were not modified. For images with decreased memorability changes were directly opposite.

Quantitatively, the difference between modified images reported in statistical values (Table 2). In addition to mean, standard deviation, and mean gradient magnitude, we also report the volume of covered gamut in CIE L*a*b* color space ($V_{gamut}$) and WLF (RSC) [40] value, which characterize perceptual contrast of an image. Given values were computed for each image excluding the oval mask and then averaged between all images of the class. Additional useful information can also be extracted from the analysis of an average histogram of the images of the same class (Fig. 6). As can be seen from the figure, poorly memorable images have a bigger number of pixels with moderate intensity values, while well-memorable images are richer on pixels with extreme values, especially dark ones. This correlates with the visually observed difference in contrast. Noteworthy is that the presented modification performed by a deep artificial network potentially can be reconstructed by a professional artist using photo editing software. Thus, in the following section, we discuss the influence of such kind of tools on image memorability.

TABLE 2. Statistical properties of generated images.

| The attribute added | mean | std | mean $|\nabla|$ | $V_{gamut}$, $10^3$ | WLF [40] |
|---|---|---|---|---|---|
| "high" | 126,0 | 58,1 | 103,9 | 67,4 | 247,5 |
| "medium" | 127,4 | 57,7 | 99,8 | 60,6 | 232,2 |
| "low" | 128,4 | 55,9 | 89,7 | 58,4 | 217,9 |

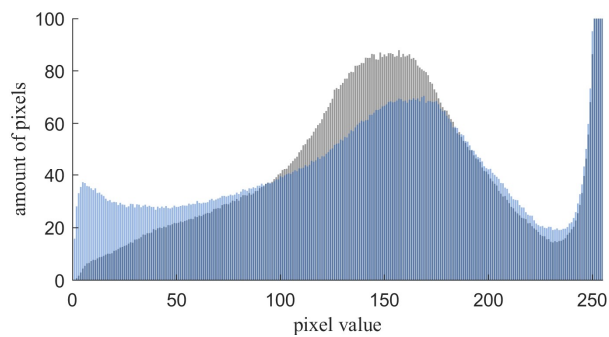

Figure 6: Mean histograms of images translated to "low" and "high" classes. black – low memorability, blue – high memorability.

## 4. Image processing for memorability modification

While the previously described method utilizes advanced deep learning algorithms for modification of images, it is also interesting to study how conventional image editing methods may influence memorability of a media. By conventional image editing methods, we denote a family of image manipulation software and raster graphics editors such as Adobe Photoshop[1], GIMP[2], Adobe Lightroom[3], Instagram[4], Snapseed[5], etc. Such software is used by designers, photographers, advertisers, and other artists to create and modify visual content. Undoubtedly, ordinary users are familiar with these tools much more than with advanced deep networks. Thus, the study of the relation between these tools and memorability is beneficial since it is closer to the real application and may be used directly.

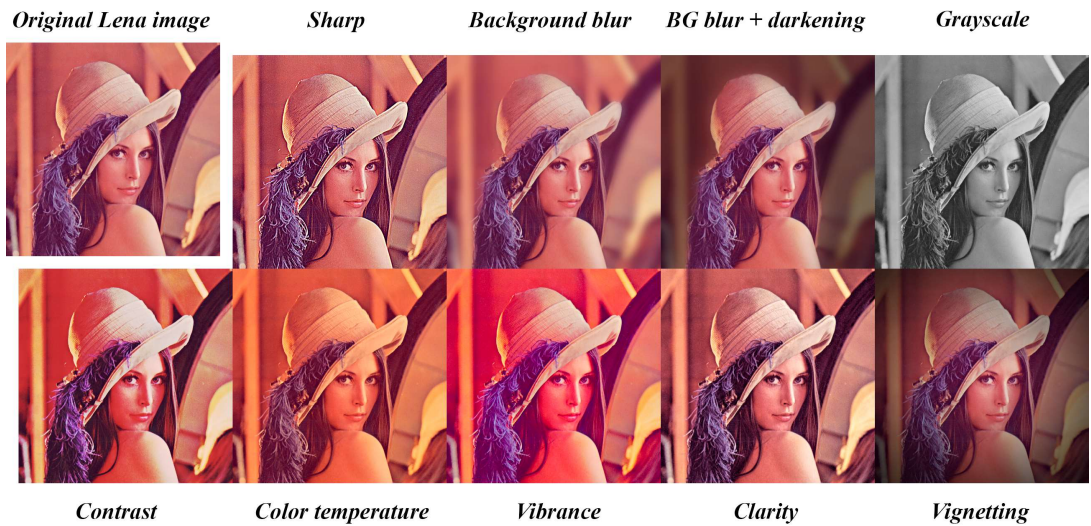

Figure 7: Set of selected image editing tools applied to Lena image.

TABLE 3. Change of memorability score (Δ) by image editing operations.

| Processing | mean | std | median | 5th % | 95th % |
|---|---|---|---|---|---|
| **Sharpening** | **+0,023** | **0,126** | **+0,015** | **-0,169** | **0,234** |
| Background blurring | -0,008 | 0,133 | -0,017 | -0,220 | 0,215 |
| **BG blurring + darkening** | **-0,049** | **0,136** | **-0,057** | **-0,261** | **0,175** |
| **Grayscaling** | **-0,023** | **0,133** | **-0,037** | **-0,244** | **0,203** |
| Contrast increasing | -0,007 | 0,131 | -0,016 | -0,209 | 0,215 |
| Color temperature 5000K | -0,002 | 0,135 | -0,009 | -0,215 | 0,220 |
| Vibrance (Saturation) | +0,005 | 0,129 | -0,004 | -0,195 | 0,220 |
| Clarity (Structure) | -0,002 | 0,133 | -0,009 | -0,211 | 0,220 |
| **Vignetting** | **-0,024** | **0,140** | **-0,029** | **-0,246** | **0,209** |

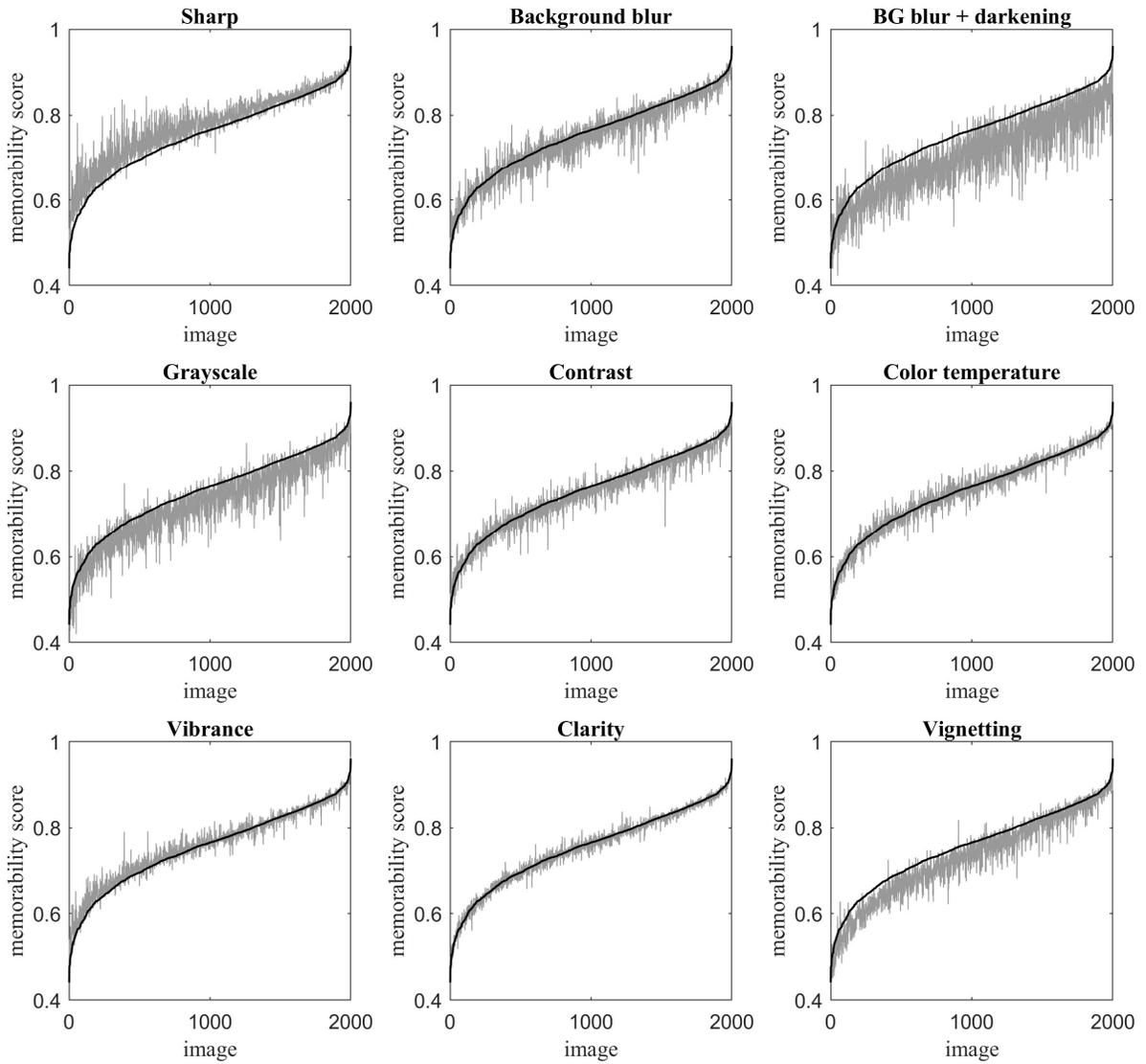

Figure 8: Influence of image editing tools on memorability. Images are sorted by original memorability score. Black line – original memorability, grey line – predicted memorability of the edited image.

4.1. Algorithms

Although each software instance utilizes unique (and confidential) algorithms, basic image enhancement techniques (e.g., contrast, saturation, smoothing, sharpening) are common for all of them. Moreover, such basic algorithms are well-known and can be reconstructed with high precision. Thus, instead of using particular software and being limited by hidden algorithms and release version, we reconstructed a number of tools in MatLab. In result, we obtained fully transparent implementation, which can be discussed, modified, and makes the experiment reproducible[6]. The set of tools selected for experiment includes: contrast increasing, Sharpening, Vibrance (Saturation) increasing, Clarity / Structure (local contrast), background blurring, Vignetting, grayscaling, color temperature shifting, and increasing volume of a photo (background blurring with darkening) (Fig. 7). Additional examples of the influence of given tools on images can be found in Appendix B.

Estimation of memorability scores was performed using pre-trained AMNet [9] (state-of-the-art memorability predictor which achieves human-like consistency). The usage of an automatic method, trained on a large set of images annotated by experimental memorability scores allowed to accurately predict an even larger number of responses for different sets of modified images, which would be impossible in a real psychophysical experiment.

4.2. Data

In this experiment, LaMem dataset [21] (the largest available memorability dataset) was used. 45,000 out of 60,000 images were used for training ("train_1" split). To remove bias, images for the experiment were taken from the test set. Moreover, in order to increase accuracy and reliability of prediction, we selected 2,000 images on which AMNet made the smallest error. Specifically, memorability scores for this 2,000 subset were predicted with an absolute error in the range from 0 to 0.02 and a mean value of 0.01, which is remarkably accurate. The selected images cover a wide range of memorability scores from 0.44 to 0.96. Considering that we tried different photo editing tools and also adjusted their parameters, given subset was modified and evaluated more than 100 times, which would not be possible with real human observers.

4.3. Results

The statistical data of memorability change after applying editing algorithms is reported in Table 3. Visualization of the results is shown in Fig. 8. The majority of studied techniques change the memorability arbitrary, providing a mean value of the memorability change close to zero. A number of methods (blur, darkening, grayscaling, and vignetting) exhibit negative influence on memorability score. We explain this effect by the loss of information (semantic and visual) caused by these methods. From another side, it may be useful to know that grayscaling of an image causes a loss in memorability equal to only -0.02 on average, which is not so dramatic comparing to the significant change of visual appearance. In rare cases basic image processing may increase memorability score, that is the case for Sharpening (it is visible especially well on Fig. 7). For a vast majority of tested images making the image sharper increased its memorability (+0.023 in average). Apart from that, Vibrance (increasing the saturation of mid-saturated colors) caused mainly an increase in memorability of poorly memorable images and a decrease in the memorability of highly memorable ones.

5. Conclusions

In this work, we proposed an innovative method of memorability modification, which works efficiently in both directions: increasing and decreasing of memorability and can produce changes of up to 33%. We also showed that basic image processing tools available in popular software like Adobe Photoshop or Instagram are not able to change memorability significantly in a predictable way. Analysis of the results confirmed that operations which lead to loss of information, such as blurring, darkening, and discoloration also lead to a drop in memorability. Sharpening is the only algorithm which showed stable enhancement of memorability score when applied to an image.

Further research may include next development: conducting additional psychophysical experiments to improve the reliability of results and collect more experimental data; application of novel GANs and other efficient machine learning algorithms to memorability modification; studying memorability maps and involving them in modification task.

---

[1] https://www.adobe.com/products/photoshop
[2] https://www.gimp.org
[3] https://lightroom.adobe.com
[4] https://www.instagram.com/about/us/
[5] https://support.google.com/snapseed
[6] https://github.com/acecreamu/changing-the-memorability

## 6. Appendix A

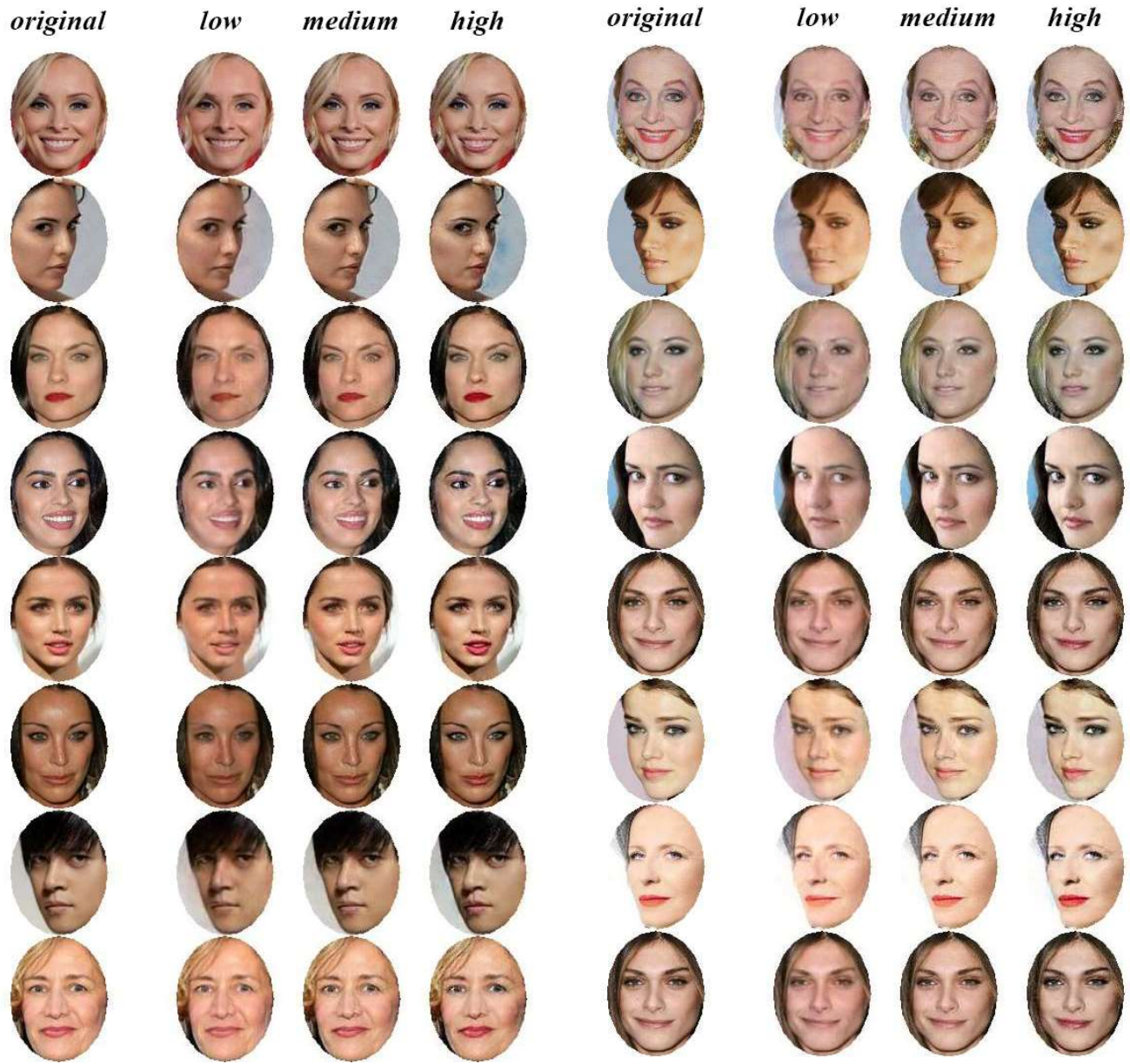

Figure 9: Additional examples of images with modified memorability.

## 7. Appendix B

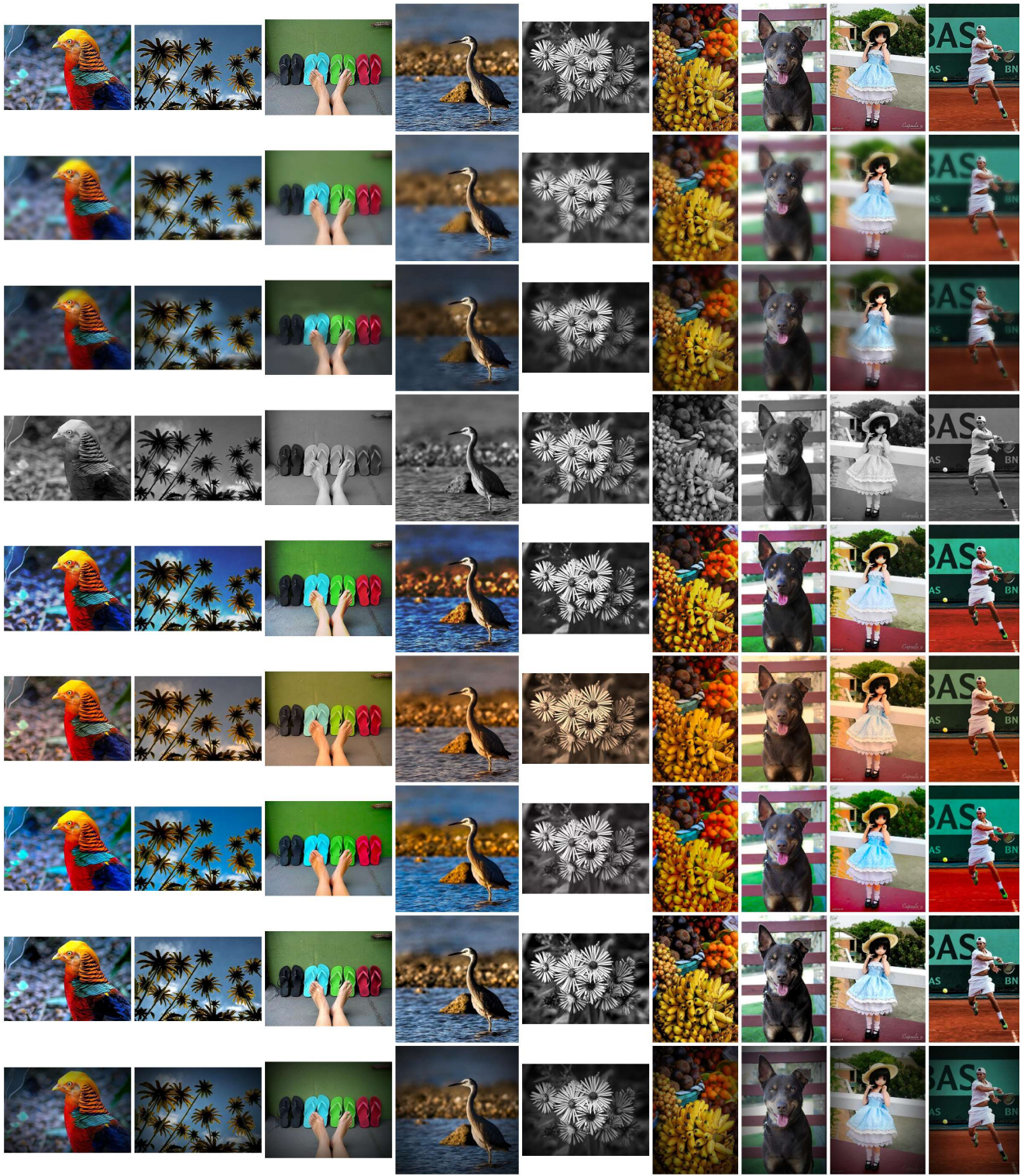

Figure 10: Demonstration of studied editing tools. Top to bottom: sharpening, BG blurring, BG blur + darkening, Grayscaling, Contrast increasing, Color temperature (5000K), Vibrance (Saturation) increasing, Clarity ("Structure", local contrast) increasing, Vignetting.